\documentclass{article}
\usepackage{spconf,amsmath,graphicx}

\usepackage{times}
\usepackage{epsfig}
\usepackage{amssymb,subfig}
\usepackage{color}
\usepackage{wrapfig}
\usepackage{gensymb}

\begin{document}
	
\title{Event Retrieval Using Motion Barcodes}
	
\name{Gil Ben-Artzi ~~~~~~~ Michael Werman ~~~~~~~ Shmuel Peleg}
		
\address{School of Computer Science and Engineering\\
        The Hebrew University of Jerusalem, Israel}
	
\maketitle
	
\begin{abstract}
We introduce a simple and effective method for retrieval of videos showing a specific event, even when the videos of that event were captured from significantly different viewpoints. Appearance-based methods fail in such cases, as appearances change with large changes of viewpoints. 

Our method is based on a pixel-based feature, ``motion barcode", which records the existence/non-existence of motion as a function of time. While appearance, motion magnitude, and motion direction can vary greatly between disparate viewpoints, the existence of motion is viewpoint invariant. Based on the motion barcode, a similarity measure is developed for videos of the same event taken from very different viewpoints. This measure is robust to occlusions common under different viewpoints, and can be computed efficiently.

Event retrieval is demonstrated using challenging videos from stationary and hand held cameras. 
\end{abstract}


\vspace{-0.1cm}
\section{Introduction}
\vspace{-0.1cm}	

Given a query video, the goal is to retrieve all other videos showing the same event at the same time. We consider cases where current appearance-based methods may fail. The first case includes videos of an event taken from significantly different viewpoints. The retrieval process can fail due to appearance changes in same event between views.  The second case includes videos of different events which take place in the same location at different times. The retrieval process can mistakenly match different events due to the similar background. Appearances-based descriptors such as SIFT \cite{lowe2004distinctive}, SURF \cite{bay2006surf} and GIST \cite{oliva2001gist} do not work well in such cases.

\begin{figure}[tb]
	\centering{
		\includegraphics[width=0.23\textwidth]{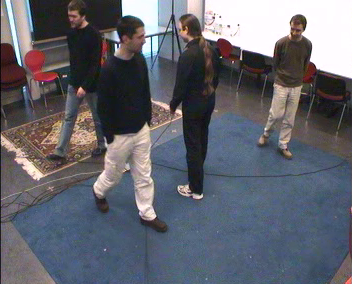}
		\includegraphics[width=0.23\textwidth]{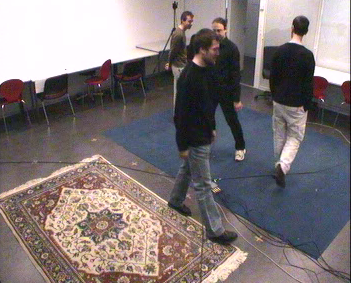}\\        
		\caption{Two significantly different views showing the same instance of the same scene. Using Motion Barcodes, more than 200 matches are found, while only 3 matches were found with SIFT descriptors.        
\label{Fig:SIFT}}}
\end{figure}

Many events are captured from multiple viewpoints. A sports event, for example, is captured by people from all around the arena. One of the key challenges is the fact that the appearance of objects in the scene depends on the viewpoint. Even the motion direction is different in each viewpoint. Fig.~\ref{Fig:SIFT} shows two views of the same instance of the same event. In the left view the actor in front is moving forward with his face visible, whereas in the right view the same actor is moving in the opposite direction and only his back is seen. In addition, in the left view the movements of all other actors can be clearly observed, where in the right view some movements are occluded. 

State-of-the-art methods \cite{revaud2013event,cao2012scene,douze2013stable} focus on video retrieval in cases where videos of events are taken from moderate viewpoint changes. They are based on representing each frame by appearance-based descriptors which are then processed into effective video representations. In \cite{revaud2013event,douze2013stable}, the video representation is the multi-VLAD descriptors \cite{jegou2010aggregating} where SIFTs are used as frame descriptors. In \cite{cao2012scene}, GISTs are the frame descriptors. The usage of appearance-based descriptors for event retrieval is due to their proven discrimination power. Motion-based descriptors which are widely used in activity recognition are not discriminative enough for such tasks.

The main contribution of this paper is the introduction of a simple and highly effective feature, the motion barcode, for event retrieval in cases where traditional appearance-based descriptors fails. The motion barcode overcomes the difference in appearance and in motion by using only the existence of motion over time. A motion barcode is computed for each pixel without incorporating  spatial information, and does not consider motion direction or magnitude. These properties make it highly invariant for different viewpoints.

    Using motion information in each pixel was proposed by Liu et al. \cite{steadyflow}. They used the motion vector of each pixel, denoted as ``pixel profile'', to stabilize the video sequence. However, the motion vector  depends on the viewing direction and therefore can not be used in our context.

	We begin by describing the motion barcodes, and continue with the similarity measure between the motion barcodes in two videos. We conclude with experiments, showing that while traditional descriptors are not sufficient our methods are successful  in the retrieval task.

\section{The Motion Barcodes}
\label{Sec:barcode}

\begin{figure}[tbf]
		\centering{
			\includegraphics[width=0.43\textwidth]{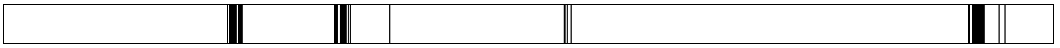}\\
			\includegraphics[width=0.43\textwidth]{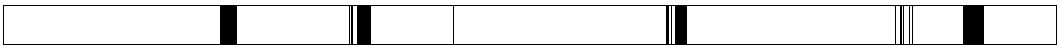}\\		
			\caption{
		Two matching motion barcodes in videos taken from two different viewing directions of the same scene. The horizontal axis is time, the black periods represent motion, and the white periods represent no motion. Matching barcodes can be $1$ (moving) in one viewpoint and $0$ (stationary) for another viewpoint. Even when the barcode is $1$ for both viewpoints, it can be due to different objects.
\label{Fig:bv}}}
\end{figure}        
     
\vspace{-0.25cm}

\noindent
{\bf Pixel-based Motion Barcode}  Corresponding pixels in different views should have the same stationary/non-stationary state at the \emph{same time} and  the longer we inspect these pixels, the more similar they will appear. 
	
	Given a sequence of $N$ video frames, the $N$ bit motion barcode $B$ of a pixel $x,y$,  
$$
	B_{x,y}(t) = \left\{
	\begin{array}{l l}
	1 &\quad \text{there is motion in pixel $(x,y)$ at time $t$}\\
	0 & \quad \text{otherwise}
	\end{array} \right.
	$$    

    We used background subtraction \cite{vibe:2011} to determine motion in a pixel. An example of two  motion barcodes of corresponding pixels in different views can be seen in Fig.~\ref{Fig:bv}.  The variations between motion barcodes are due to the fact that each reflects the motion along of a $3D$ ray. Thus, each of the above motion barcodes include  movements that are not be observed by the other.\\ 

\vspace{-0.25cm}

\begin{figure}[t]
		\centering{
			\includegraphics[width=0.48\textwidth]{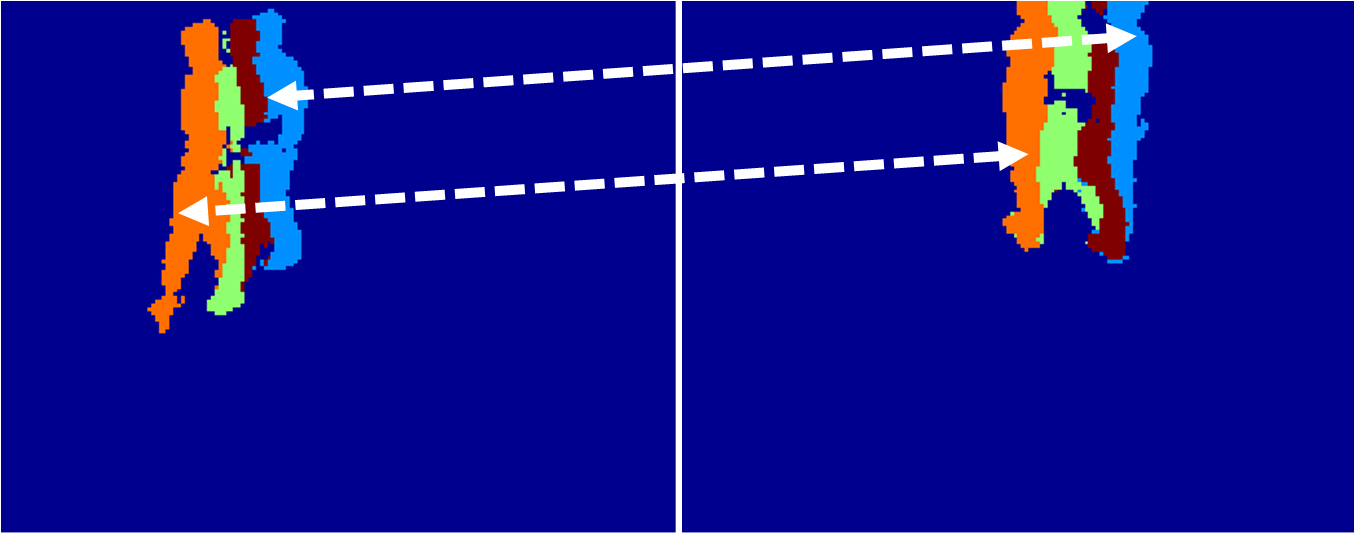}	
						
			\caption{Two views of the same scene, with corresponding motion barcodes. In the left image there is an actor walking from back to front and in the right image the same actor is walking from front to back. The two cameras have 90 degrees difference in viewing directions. To enable visibility, motion barcodes were reduced to 5 clusters by k-means. Each of the five clusters has a different color. Each region of pixels in the left view corresponds to a region of pixels in the right view.  The motion barcodes in corresponding regions have the highest correlation score. The arrows are examples for such two regions. 
	\label{Fig:kmeans}}}
\end{figure}

\noindent
{\bf Similarity between Motion Barcodes}. The similarity between two motion barcodes is their correlation. As an example, the motion barcodes in Fig.~\ref{Fig:bv} are highly correlated even though  there are additional movements in each view. Fig.~\ref{Fig:kmeans} shows two views of the same event with a 90\degree \,  difference in viewing directions. The pixels were clustered according to their motion barcodes. The motion barcodes in each region at the left view has the highest correlation score with the motion barcodes in the corresponding region in the right view.\\

	\begin{figure}[tb]
		\centering{
            \includegraphics[width=0.3\textwidth]{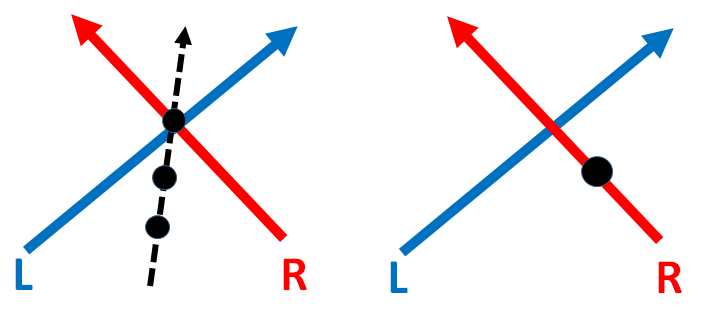}
			\includegraphics[width=0.16\textwidth]{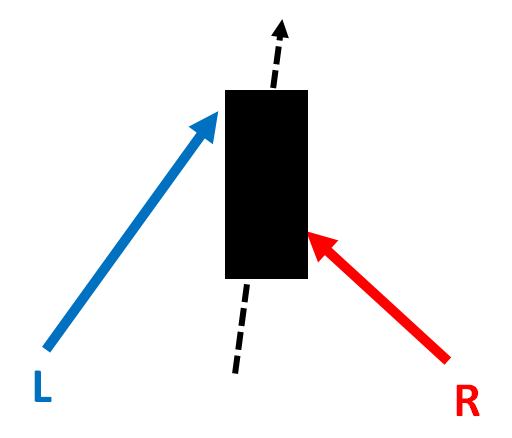}\\
			(a)~~~~~~~~~~~~~~~~~~~~~~~~(b)~~~~~~~~~~~~~~~~~~~~~~~(c)\\
						
			\caption{ (a) The ideal case where a point in the scene is moving across the intersection of two rays. L is for a ray (pixel) from the left camera, R is for a ray (pixel) from the right camera. Both pixels observe the existence of motion simultaneously.  (b) Same pixels (rays) as in (a), but motion is seen only in the right camera. (c) Due to the 3D volume of objects in the scene, pixels can be matched even when they observe different points on an object. 
	\label{Fig:rays}}}
	\end{figure}

\vspace{-0.25cm}

\begin{figure}[tb]
		\centering{
			\includegraphics[width=0.215\textwidth]{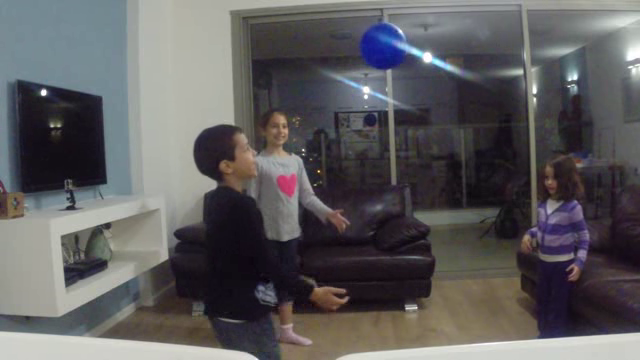}
			\includegraphics[width=0.22\textwidth]{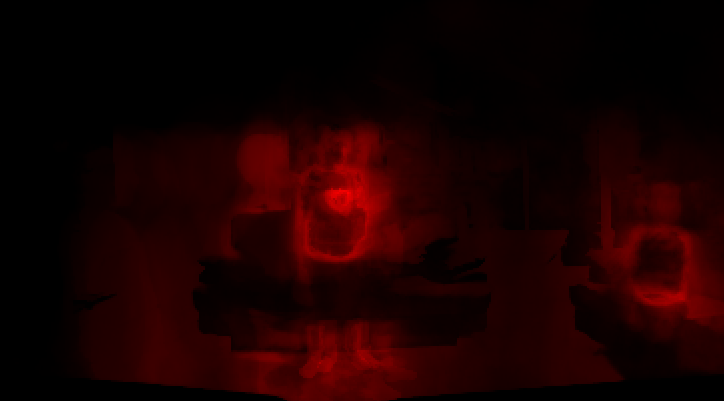}\\
            (a)~~~~~~~~~~~~~~~~~~~~~~~~~~~~~~~~~~(b)\\
			\includegraphics[width=0.22\textwidth]{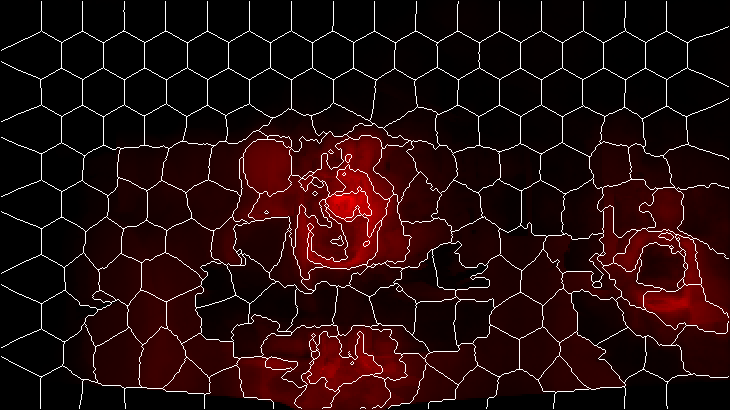}
            \includegraphics[width=0.22\textwidth]{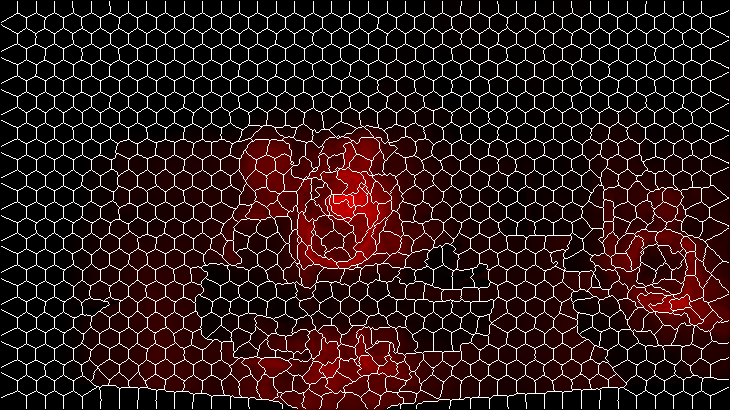}\\
			(c)~~~~~~~~~~~~~~~~~~~~~~~~~~~~~~~~~~(d)\\
			
			
			\caption{Selecting one motion barcode per superpixel. (a) The original scene. (b) The motion image, where each pixel is the sum of ``1"s in its motion barcode. (c) The  segmentation of the motion picture to superpixels using the SLIC algorithm, with 200 regions (d) The superpixel segmentation with 1000 regions.
				\label{Fig:sp}}}
	\end{figure}

\noindent
{\bf Pooling}. In a typical sequence there are more than 300K non-zero motion barcodes, many of them representing similar motions. Fig.~\ref{Fig:rays} shows the matching pixels, as $3D$ rays, based on the existence of such motion. Since many points in an object move together, matching is possible even between image points that are not projections of same 3D point. We pool the motion barcodes by segmenting the video into superpixels and select a single barcode for each superpixel.  Superpixels are computed from a ``motion image" $M$, where each pixel in $M$ is the number of ``1"s in its motion barcode, $ M(x,y)= \sum_{t}B_{x,y}(t)$. Segmentation to superpixels is performed on the motion image $M$ using the SLIC algorithm \cite{achanta2012slic} as can be seen Fig.~\ref{Fig:sp}. For each superpixel we chose a representative barcode as the rounded average of all its barcodes. This  barcode minimizes the sum of hamming distances to all other motion barcodes in the superpixel.

Pooling the motion barcodes into superpixels yields a fixed size representation  and makes this representation robust and effective. A similar approach was introduced in \cite{taralova2014motion} for a fixed size video representation, by pooling features in video into supervoxels.\\

\vspace{-0.3cm}

\noindent
{\bf Similarity between videos}. The similarity score between the two videos can be evaluated based on the optimal assignment using the bipartite matching algorithm \cite{kuhn1955hungarian}. Let $B^i_{1\ldots K_i}$ be the motion barcodes of Clip $i$.  in the bipartite matching the weight between $B^1_i$ and $B^2_j$ is their correlation. 
In practice, we use the following heuristic, with very similar results, running  $100 \,\times$ faster.

\begin{figure}[b]
	\centering{
		\includegraphics[width=0.23\textwidth]{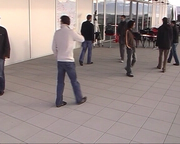}
		\includegraphics[width=0.23\textwidth]{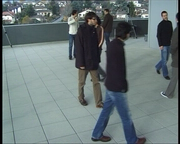}\\        
		\caption{An example of two different views of the same scene from the EPFL multi-view dataset.       
\label{Fig:Dataset}}}
\end{figure}

Let $C^i$ be the number barcodes $B^i_j$ having at least one match at the other video clip with correlation higher than a given threshold. We normally use a threshold of $0.4$. A discussion about the effect of this threshold appears in Sec.~\ref{Sec:Results}.
The similarity between the two video clips is $\frac{C^1}{K_1}+\frac{C^2}{K_2}$.
The similarity is the fraction of vertices in each video having at least one above-threshold match in the other video clip. The higher the similarity score, the more similar the videos are. The threshold values will be discussed  in Sec.~\ref{Sec:Results}. 

The similarity measure  we  used is similar to the bag-of-words model with binary weights. The key difference is that in our case we do not have a common codebook  shared between all the sequences. 

\section{Experiments}
\label{Sec:Results}

\vspace{-0.2cm}

\noindent
{\bf Dataset}. Standard event retrieval datasets, such as EVVE \cite{evve}, are based on events in which appearance-based descriptors are useful. In order to reflect the challenges we consider, we used the EPFL Multi-camera \emph{pedestrians} dataset \cite{Berclaz11,Fleuret08a}. It includes 30 sequences of 6 different scenes under significant multi-view settings, some of them take place in the same location but in different times. The scenes are both indoor and outdoor and include many occlusions. An example can be seen in  Fig.~\ref{Fig:Dataset}\\

\begin{table}[tb]
\begin{center}
    \begin{tabular}{  r  c  c } 
       ~~~  & Same Event & Different Event  \\ \hline
    Same Viewpoints & 170 & 110 \\ 
    Different Viewpoints & 1.9 & 1.8 \\
	\end{tabular}
 \end{center}
 
 \caption{Matching frames using appearance-based descriptors. The results are the mean number of SIFT descriptors matched across different video sequences. On average, comparing a sequence to a time shifted version of itself matched 170 descriptors. For significantly different viewing directions of the same event there were on average 1.9 matching descriptors. 
 \label{Fig:dDirections}} 
\end{table}

\noindent
{\bf Appearance-based Descriptors}. In order to evaluate the performance we compared the mean AP to appearance-based state-of-the art methods. Following \cite{revaud2013event,douze2013stable}, we used SIFT+VLAD as our frame descriptors. As a baseline, we  also used SIFT+BoW. The codebook was created as in \cite{revaud2013event} and the distances between the clips evaluated accordingly.
Table~\ref{Fig:dDirections} shows the average number of SIFTs matched 
between the sequences. For the same event, there were on average only 1.9 matching descriptors in more than 90\% of the significantly different viewing directions. 
Moreover, there are many false matches between \emph{different} events due to the fact that they share the same background.  It can be seen that when the angle between the cameras is too wide or when the event is taking place in the same location (e.g. stadium) but at a different time, such descriptors completely fail.\\

\begin{figure}[tb]
	\centering{		
		\includegraphics[width=0.235\textwidth]{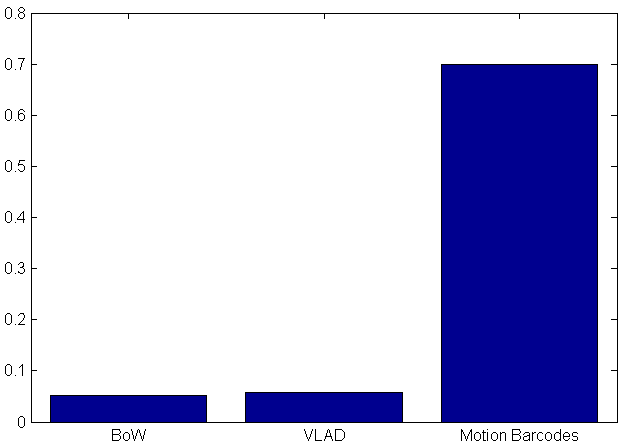}
        \includegraphics[width=0.215\textwidth]{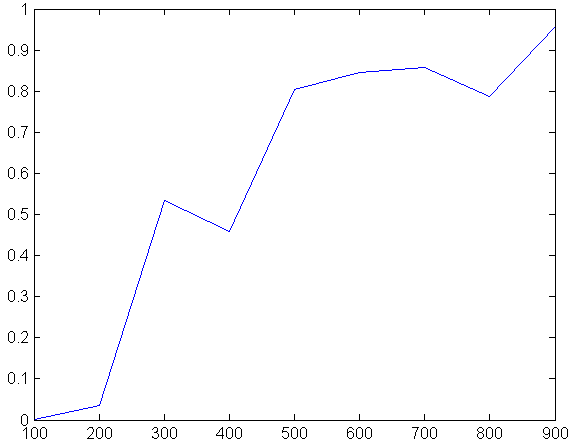}\\ 
\caption{ (a) The mean average precision (mean AP) of SIFT+BoW, SIFT+VLAD and Motion Barcodes for event retrieval task. We can see that the motion barcode is highly effective. (b) The mean AP as a function of the number of motion barcodes with sufficient motion. The more motion barcodes with enough motion, the higher accuracy we have. 
\label{Fig:meanAP}}}
\end{figure}

\begin{figure}[tbh]
	\centering{		
		\includegraphics[width=0.23\textwidth]{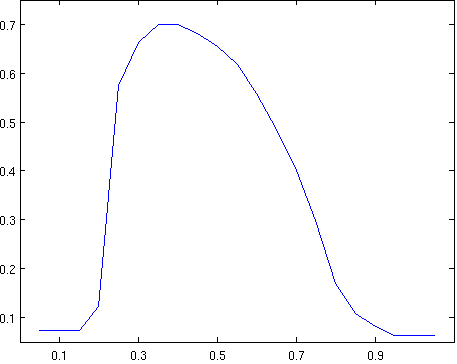}
        \includegraphics[width=0.235\textwidth]{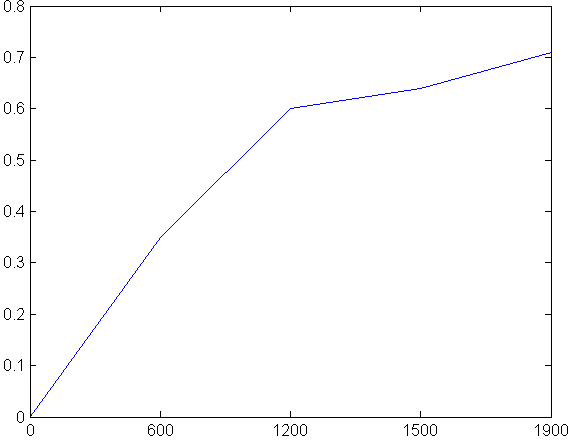}\\        
\caption{ (a) The mean AP as a function of the similarity correlation threshold. The y-axis is the mean AP. The x-axis is the correlation threshold. 
(b) The mean AP as a function of the length of the motion barcodes used. 
\label{Fig:meanAP1}}}
\end{figure}

\noindent
{\bf Evaluation method}. The dataset was divided into 200 different clips. We  added 200 distractor clips, where the distractors are with similar activities  (e.g. walking).  Every clip from the dataset was used as a query and the rest of the clips within the set were used as the target database.  For each clip, there were  2 or 3 true matches. The mean average precision (mean AP) was evaluated over all clips in all sets. 

To use the motion barcodes in the retrieval process we removed non-informative barcodes from each clip, the motion barcodes that did not have enough motion. We required that the motion (1's) in each motion barcode  be more than 10\% of its length. We also required a minimal number of motion barcodes, empirically set to 100. The effect of this requirement  will be discussed later. \\

\noindent
{\bf Results}. Fig.~\ref{Fig:meanAP}.a presents the effectiveness of the motion barcodes, under such extreme cases. It shows the performance of each of the methods in the retrieval process. The motion barcode outperformed both BoW and VLAD, with mean AP of 0.7 vs. 0.051 and 0.058 respectively.  Fig.~\ref{Fig:meanAP}.b shows the  accuracy as a function of the number of barcodes with sufficient motion. For sequences with sufficient motion the accuracy is high and for sequences with not enough motion the accuracy is lower. It is therefore expected that for events with sufficient motion the mean AP will be very high. We can also empirically predict beforehand the ability to successfully retrieve an event by the number of motion barcodes with sufficient motion it has.

Fig.~\ref{Fig:meanAP1}.a shows the effect of the correlation threshold used for matching  barcodes on the mean AP. We can see that there is a trade-off between the specificity we require and the robustness to the viewing directions.  The best result is reached in a correlation threshold of $0.4$ with peak mean AP is $0.7$. It can also be seen that the similarity is robust to the correlation threshold since small fluctuations almost have no effect on the mean AP. Fig.~\ref{Fig:meanAP1}.b shows that the longer the motion barcodes are, the more accurate the similarity.
The peak mean AP for motion barcodes was found when using motion barcodes of length 1000. This was found by evaluating the mean AP for different number of superpixels.\\

	\begin{figure}[tb]
		\centering{
			\includegraphics[width=0.22\textwidth]{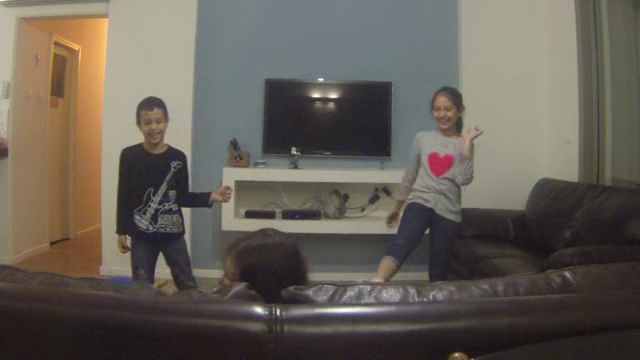}
			\includegraphics[width=0.22\textwidth]{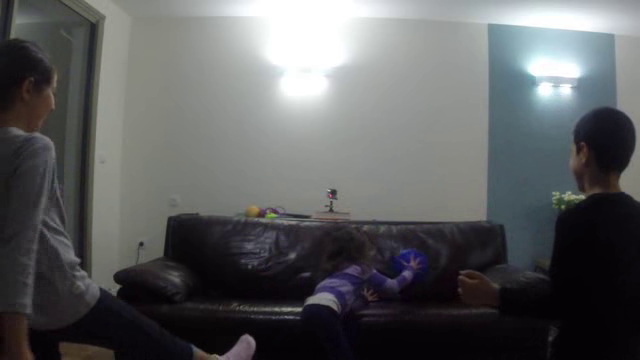}\\
            (a)~~~~~~~~~~~~~~~~~~~~~~~~~~~~~~~~~~~(b)\\
			\includegraphics[width=0.22\textwidth]{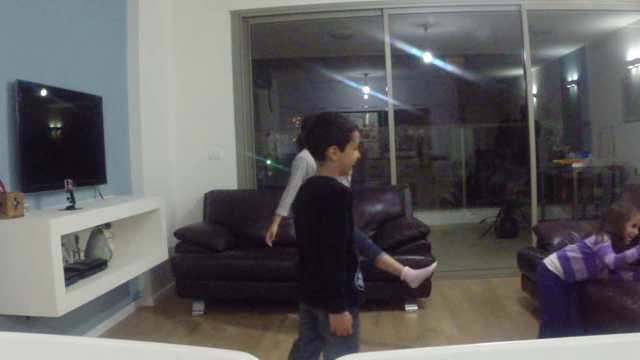}
            \includegraphics[width=0.22\textwidth]{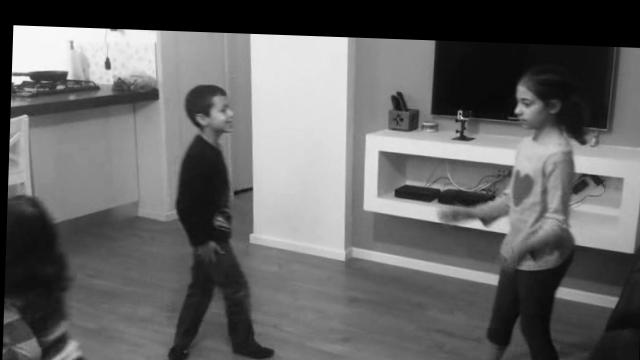}\\
			(c)~~~~~~~~~~~~~~~~~~~~~~~~~~~~~~~~~~~(d)\\
            \includegraphics[width=0.22\textwidth]{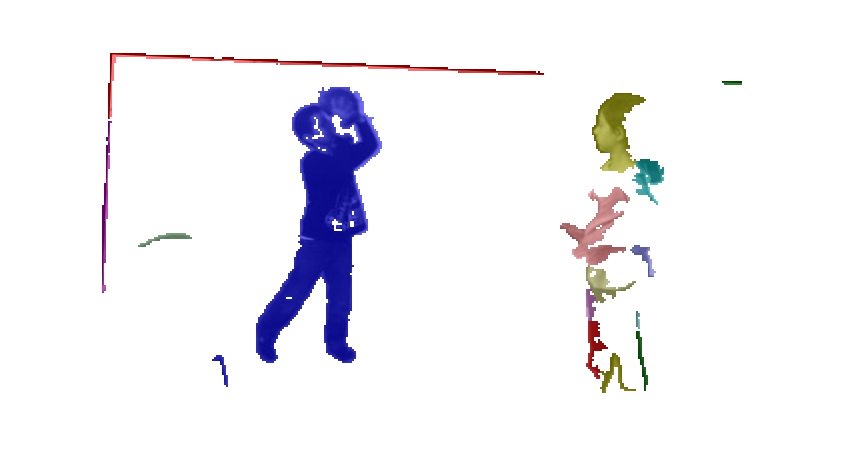}
			\includegraphics[width=0.22\textwidth]{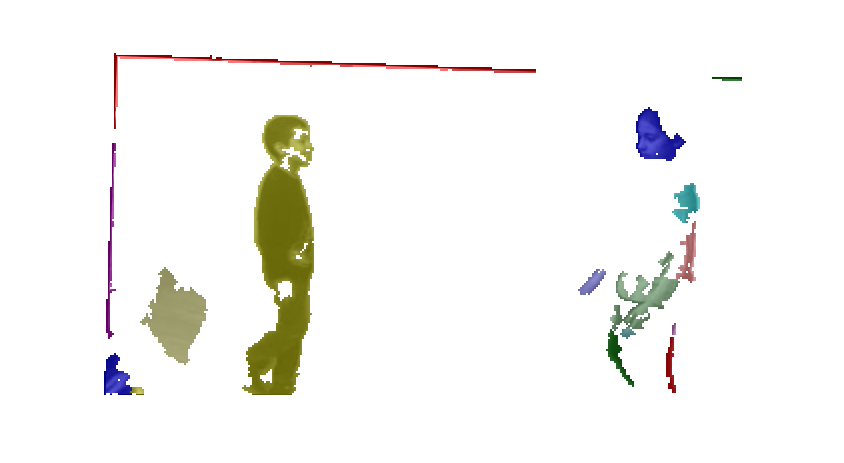}
			(e)~~~~~~~~~~~~~~~~~~~~~~~~~~~~~~~~~~~(f)\\
            
			\caption{Four simultaneous frames from 4 very different views. (a)-(c) are taken by  stationary 
cameras, and (d) is taken by a hand-held mobile phone. (e)-(f) Motion detection masks on two stabilized frames from the mobile phone. Although there are many motion detection errors, the motion barcodes are sufficient to recover the  event. 
				\label{Fig:distortion1}}}
	\end{figure}
    
\noindent
{\bf Stabilized Sequences}. In order to verify our results on hand held cameras we captured an event from 3 stationary cameras and one hand held mobile phone, a total of 4 views. The mobile phone sequence is with significant shakes, and was stabilized using homographies \cite{Matsushita05full-framevideo}. 
Fig.~\ref{Fig:distortion1} shows frames taken at same time from 3 stationary views and a mobile phone. Fig.~\ref{Fig:distortion1}.e-f show the motion detection masks on two stabilized frames of the mobile phone. It can be seen that the motion detection masks are not perfect.  All sequences from this setup were tested with the existing dataset and we compared the results with and without the stabilized sequence. The results of the stabilized sequence is similar to the stationary sequences with peak mean AP of 0.68. 

\vspace{-0.3cm}

\section{Concluding Remarks}

\vspace{-0.4cm}

We introduced motion barcodes, a robust motion feature that can be used to determine if two videos show the same event even when the videos were taken from very different viewing directions. In such cases appearance based methods may fail, as well as traditional motion-based features. We propose a similarity measure between motion barcodes, and show its effectiveness for event retrieval tasks in challenging settings.

\vspace{0.2cm}
    
\noindent\textbf{Acknowledgment:} This research was supported by Google, by Intel ICRI-CI, and by Israel Science Foundation.

	{\small
		\bibliographystyle{IEEEbib}
		\bibliography{refs}

\begin{thebibliography}{10}

\bibitem{lowe2004distinctive}
David~G Lowe,
\newblock ``Distinctive image features from scale-invariant keypoints,''
\newblock {\em IJCV}, vol. 60, no. 2, pp. 91--110, 2004.

\bibitem{bay2006surf}
Herbert Bay, Tinne Tuytelaars, and Luc Van~Gool,
\newblock ``Surf: Speeded up robust features,''
\newblock in {\em ECCV'06}, pp. 404--417. 2006.

\bibitem{oliva2001gist}
Aude Oliva and Antonio Torralba,
\newblock ``Modeling the shape of the scene: A holistic representation of the
  spatial envelope,''
\newblock {\em IJCV}, vol. 42, no. 3, pp. 145--175, 2001.

\bibitem{revaud2013event}
J{\'e}r{\^o}me Revaud, Matthijs Douze, Cordelia Schmid, and Herv{\'e}
  J{\'e}gou,
\newblock ``Event retrieval in large video collections with circulant temporal
  encoding,''
\newblock in {\em Computer Vision and Pattern Recognition (CVPR), 2013 IEEE
  Conference on}. IEEE, 2013, pp. 2459--2466.

\bibitem{cao2012scene}
Liangliang Cao, Yadong Mu, Apostol Natsev, Shih-Fu Chang, Gang Hua, and John~R
  Smith,
\newblock ``Scene aligned pooling for complex video recognition,''
\newblock in {\em Computer Vision--ECCV 2012}, pp. 688--701. Springer, 2012.

\bibitem{douze2013stable}
Matthijs Douze, J{\'e}r{\^o}me Revaud, Cordelia Schmid, and Herv{\'e}
  J{\'e}gou,
\newblock ``Stable hyper-pooling and query expansion for event detection,''
\newblock in {\em Computer Vision (ICCV), 2013 IEEE International Conference
  on}. IEEE, 2013, pp. 1825--1832.

\bibitem{jegou2010aggregating}
Herv{\'e} J{\'e}gou, Matthijs Douze, Cordelia Schmid, and Patrick P{\'e}rez,
\newblock ``Aggregating local descriptors into a compact image
  representation,''
\newblock in {\em Computer Vision and Pattern Recognition (CVPR), 2010 IEEE
  Conference on}. IEEE, 2010, pp. 3304--3311.

\bibitem{steadyflow}
Shuaicheng Liu, Lu~Yuan, Ping Tan, and Jian Sun,
\newblock ``Steadyflow: Spatially smooth optical flow for video
  stabilization,''
\newblock in {\em Computer Vision and Pattern Recognition (CVPR), 2014 IEEE
  Conference on}, June 2014, pp. 4209--4216.

\bibitem{vibe:2011}
O.~Barnich and M.~Van~Droogenbroeck,
\newblock ``Vibe: A universal background subtraction algorithm for video
  sequences,''
\newblock {\em IEEE Transactions on Image Processing}, vol. 20, no. 6, pp.
  1709--1724, 2011.

\bibitem{achanta2012slic}
Radhakrishna Achanta, Appu Shaji, Kevin Smith, Aurelien Lucchi, Pascal Fua, and
  Sabine Susstrunk,
\newblock ``Slic superpixels compared to state-of-the-art superpixel methods,''
\newblock {\em Pattern Analysis and Machine Intelligence, IEEE Transactions
  on}, vol. 34, no. 11, pp. 2274--2282, 2012.

\bibitem{taralova2014motion}
Ekaterina~H Taralova, Fernando De~la Torre, and Martial Hebert,
\newblock ``Motion words for videos,''
\newblock in {\em Computer Vision--ECCV 2014}, pp. 725--740. Springer, 2014.

\bibitem{kuhn1955hungarian}
Harold~W Kuhn,
\newblock ``The hungarian method for the assignment problem,''
\newblock {\em Naval research logistics quarterly}, vol. 2, no. 1-2, pp.
  83--97, 1955.

\bibitem{evve}
EVVE,
\newblock ``http://pascal.inrialpes.fr/data/evve/,'' 2013.

\bibitem{Berclaz11}
J.~Berclaz, F.~Fleuret, E.~Turetken, and P.~Fua,
\newblock ``Multiple object tracking using k-shortest paths optimization,''
\newblock {\em IEEE Transactions on Pattern Analysis and Machine Intelligence},
  2011.

\bibitem{Fleuret08a}
F.~Fleuret, J.~Berclaz, R.~Lengagne, and P.~Fua,
\newblock ``Multi-camera people tracking with a probabilistic occupancy map,''
\newblock {\em IEEE Transactions on Pattern Analysis and Machine Intelligence},
  vol. 30, no. 2, pp. 267--282, February 2008.

\bibitem{Matsushita05full-framevideo}
Yasuyuki Matsushita, Eyal Ofek, Xiaoou Tang, and Heung yeung Shum,
\newblock ``Full-frame video stabilization,''
\newblock in {\em CVPR'05}, 2005, pp. 50--57.

\end{thebibliography}
	}
\end{document}